\documentclass[conference]{IEEEtran}  
\usepackage[numbers]{natbib}
\usepackage{multicol}

\usepackage{bm}
\usepackage{times}
\usepackage{epsfig}
\usepackage{amssymb}
\usepackage{algpseudocode}
\usepackage{algorithm}
\usepackage{amsmath}
\usepackage{subfigure}
\usepackage{multirow}
\usepackage{multicol}
\usepackage{caption}
\usepackage{gensymb}
\usepackage{adjustbox}
\usepackage{mathtools}
\usepackage{booktabs}
\usepackage{pifont}
\usepackage{xspace}
\usepackage{tabularx} 
\usepackage{multirow}
\usepackage{enumerate}
\usepackage{xcolor}
\usepackage{url}
\usepackage[normalem]{ulem}
\usepackage{hyperref}
\useunder{\uline}{\ul}{}

\newcommand{\trsp}{{\scriptscriptstyle\top}}

\title{{Configuration Space Distance Fields\\for Manipulation Planning}
}
\author{Yiming Li\textsuperscript{1,2}, Xuemin Chi\textsuperscript{1,3}, Amirreza Razmjoo\textsuperscript{1,2}, and Sylvain Calinon\textsuperscript{1,2}
\\
\textsuperscript{1}Idiap Research Institute \ \ \ \ \ \ \textsuperscript{2}EPFL 
\ \ \ \ \ \  \textsuperscript{3}Zhejiang University
}

\begin{document}

\maketitle
\begin{abstract}
    The signed distance field (SDF) is a popular implicit shape representation in robotics, providing geometric information about objects and obstacles in a form that can easily be combined with control, optimization and learning techniques. Most often, SDFs are used to represent distances in task space, which corresponds to the familiar notion of distances that we perceive in our 3D world. However, SDFs can mathematically be used in other spaces, including robot configuration spaces. For a robot manipulator, this configuration space typically corresponds to the joint angles for each articulation of the robot. While it is customary in robot planning to express which portions of the configuration space are free from collision with obstacles, it is less common to think of this information as a distance field in the configuration space. In this paper, we demonstrate the potential of considering SDFs in the robot configuration space for optimization, which we call the configuration space distance field (or CDF for short). Similarly to the use of SDF in task space, CDF provides an efficient joint angle distance query and direct access to the derivatives (joint angle velocity). Most approaches split the overall computation with one part in task space followed by one part in configuration space (evaluating distances in task space and then computing actions with inverse kinematics). Instead, CDF allows the implicit structure to be leveraged by control, optimization, and learning problems in a unified manner. In particular, we propose an efficient algorithm to compute and fuse CDFs that can be generalized to arbitrary scenes. A corresponding neural CDF representation using multilayer perceptrons (MLPs) is also presented to obtain a compact and continuous representation while improving computation efficiency. We demonstrate the effectiveness of CDF with planar obstacle avoidance examples and with a 7-axis Franka robot in inverse kinematics and manipulation planning tasks. Project page: \href{https://sites.google.com/view/cdfmp/home}{\textcolor{blue}{https://sites.google.com/view/cdfmp/home}}

\end{abstract}

\section{Introduction}

\begin{table*}[t]
    \begin{center}
        \begin{tabular}{c|c}
            \includegraphics[width = 0.48\linewidth]{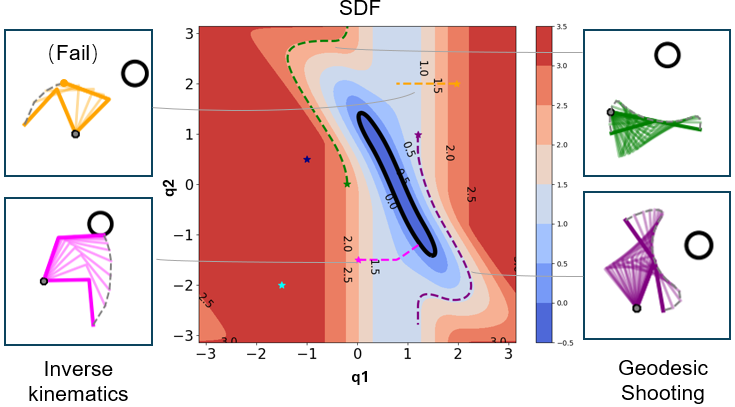} &
            \includegraphics[width = 0.48\linewidth]{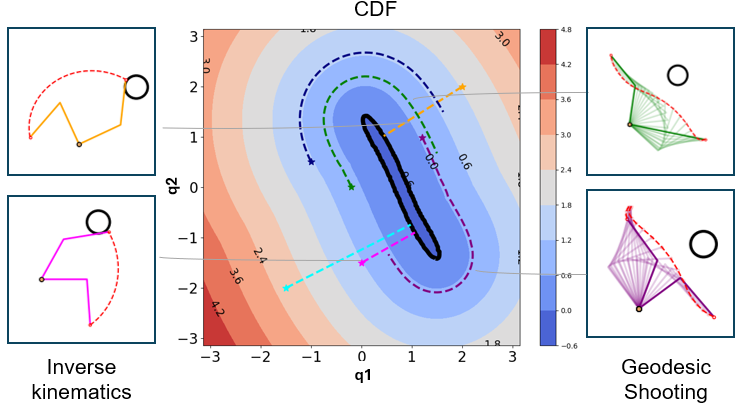} \\
            (a)&(b)\\
        \end{tabular}
        \captionof{figure}{Differences between SDF and CDF. The colored level sets depict the distances to the object, where the black contours represent joint angles leading to robot-object surface contact. The zero-level-set of SDF and CDF is the same, but the other level sets of CDF are characterized by evenly expanded distances and unit norm gradient. This property leads to gradient projection that can directly be computed, which solves the inverse kinematics problem for the contact task in one step (see trajectories in cyan and pink), whereas SDF requires multiple iterations while encountering singularities, which can even fail due to gradient vanishing (see trajectory in yellow). \textcolor{black}{Geodesics on the CDF naturally wrap around the shape of the object in configuration space (see trajectories in blue, green and purple).}
        }
        \label{fig:overview}
        \vspace{-6mm}
    \end{center}
\end{table*}
 
Distances are the most fundamental and intuitive metrics for expressing the interrelation among multiple variables. In robotics, they are typically used to measure the geometric relationship among diverse representations, such as points, poses, trajectories, surfaces and shapes, which are exploited in various tasks including inverse kinematics~\cite{nakamura1986inverse, chiacchio1991closed} and manipulation planning~\cite{lynch2017modern}. The signed distance field (SDF) representation has become a popular representation, that can for example be used to encode the Euclidean distance from a point to an object boundary. The differentiability and unit norm gradient properties make it easy to integrate into learning~\cite{weng2023neural, driess2022learning,sutanto2021learning}, optimization~\cite{ratliff2009chomp,ratliff2015understanding, sundaralingam2023curobo}, and control~\cite{curless1996volumetric,quiroz2019whole,liu2022regularized}.

SDFs are conventionally employed in 3D task spaces. In the context of manipulation, a typical control task is composed of two steps, first using an SDF in task space to evaluate the distance to the object, followed by an inverse kinematics step to find the joint angle configuration that can reduce this distance. Because of the nonlinear mapping between task space and joint space, this problem is typically solved with a few iterations by second-order optimization (corresponding to the use of a Jacobian pseudoinverse at each iteration).   

Figure~\ref{fig:overview}-\emph{left} depicts this process visualized in the configuration space, where the full distance field has been computed to be depicted as colored level sets (in practice, we evaluate the forward kinematics function and associated Jacobian only at the current joint configuration of the robot).

While conventionally employed in 3D task spaces, SDFs can be considered in other spaces, including robot joint configuration spaces in which planning and control problems take place. We see in Figure~\ref{fig:overview}-\emph{left} that when we transpose an SDF from task space to configuration space, the property of unit norm gradient disappears, while in Figure~\ref{fig:overview}-\emph{right}, this property is maintained. 
We will refer to this approach as a Configuration Space Distance Field (CDF), a scalar field measuring the angular distance between joint angles and the object geometry in configuration space (Fig.~\ref{fig:overview}-\emph{right}). For manipulation tasks, CDF directly estimates the minimum joint motion required by the robot to establish contact with an object, with gradients consistently pointing toward the object. Unlike SDFs in task space, CDF is directly formulated in configuration space, preserving the Euclidean property of the distance field, ensuring a uniform span of distances and maintaining unit magnitudes in gradient directions.

CDF offers several advantages. It naturally bridges task space and configuration space, providing a unified approach to solving problems that traditionally involve computation in separate spaces. For instance, the inverse kinematics problem is usually solved by evaluating the task space distance and then computing joint space motions. In contrast, CDF solves this problem through one-step gradient projections, avoiding Gauss-Newton iterations. It also offers intuitive geodesics that reflect object geometry in configuration space, see Fig.~\ref{fig:overview}. Furthermore, CDF inherits the merits of SDFs, including implicit structure, Boolean operations for composition involving multiple SDFs, efficient queries, and differentiability. We also propose a neural variant called neural CDF. Analogously to neural SDFs, neural CDFs also offer a compact, continuous, analytical, and latent space representation, thereby facilitating seamless integration into learning, optimization, and control frameworks.
Most approaches developed for SDFs can be directly applied to CDF, which directly solves problems in configuration space.

The organization of this paper is as follows. Section~\ref{sec:related_work} reviews the related work. Section~\ref{sec:CDF} discusses the formulation, properties, computation and fusion strategy of CDF. Section~\ref{sec:lCDF} presents the implicit neural CDF representation. Section~\ref{sec:wik} and~\ref{sec:mp} demonstrate the effectiveness of CDF in whole-body inverse kinematics and manipulation planning tasks. Section~\ref{sec:conclusion} discusses and concludes the paper. The primary contributions of this paper are:

\begin{itemize}
    \item The introduction of the CDF representation, offering a unified framework for addressing robot manipulation challenges within configuration space.
    \item An efficient algorithm for calculating the zero-level-set geometry of the object in configuration space, corresponding to the set of robot configurations that will lead to a collision. A subsequent fusion strategy is also proposed to combine multiple CDFs online, enabling the generalization of CDF to diverse scenes.
    \item A neural CDF variant utilizing a multilayer perceptron (MLP), together with the design of the corresponding loss functions, resulting in a concise and continuous representation. This variation provides trade-offs across efficiency, accuracy, and compression capabilities while keeping a simple and flexible structure. 
    \item Experiments comparing CDF with SDFs and their derivatives on a planar robot and a 7-axis Franka robot. The conducted experiments highlight the efficiency of CDF in solving inverse kinematics tasks through gradient projection and its effectiveness in addressing manipulation planning challenges, leading to the generation of natural robotic motions.
\end{itemize}

\section{Related Work}
\label{sec:related_work}

SDFs have garnered extensive attention in the field of computer version and graphics, particularly in shape encoding~\cite{park2019deepsdf,gropp2020implicit}, mesh generation~\cite{remelli2020meshsdf}, and differentiable rendering~\cite{vicini2022differentiable,mildenhall2021nerf}. Their efficacy in distance and gradient queries has made them popular in robotics, with applications in planning~\cite{ratliff2009chomp,zucker2013chomp}, mapping~\cite{izadi2011kinectfusion}, and manipulation~\cite{driess2022learning}. Recent studies proposed to encode SDF with joint angles~\cite{koptev2022neural}, or learn an SDF representation of the swept volume of robot manipulators~\cite{michaux2023reachability}, to model the SDF of articulated robots and apply it to manipulation tasks~\cite{li2024representing}.

The prevailing focus on SDFs in robotics centers on task space, where configuration space actions are typically computed independently through mappings between the two spaces~\cite{Siciliano2008robotics,ratliff2018riemannian}. Existing approaches often model the configuration space using binary maps denoting collision status of joint configurations~\cite{schulman2014motion,werner2023approximating}, to support sample-based motion planning algorithms~\cite{williams2017model,bhardwaj2022storm,jankowski2023vp}. Despite significant progress, these control and planning strategies are computationally expensive in high dimensional space due to the lack of gradient information. 

In contrast, considering a distance field in the configuration space introduces new control and planning strategies, by shifting the focus from conventional binary collision masks to continuous and structured representations. For instance, with this approach, the inverse kinematics problem simplifies to an SDF pullback in configuration space, requiring only one-step gradient projection. More generally, CDF enables the transposition of SDF methodologies developed for task space to configuration space. The computation can be viewed as a point-mass system, while obstacles form topological holes in configuration space and geodesics produce natural curved paths around them~\cite{ratliff2015understanding}. The approach can be extended to geometric motion planning frameworks, including Riemannian motion policies~\cite{ratliff2018riemannian}, geometric fabrics~\cite{van2022geometric}, and dynamic-aware motion optimization by assigning metrics on the distance field~\cite{klein2023design,RCFS}.

\section{Configuration Space Distance Field (CDF)}
\label{sec:CDF}

\begin{algorithm}[t]
    \caption{Finding 0 level-set configurations}
    \label{algorithm1}
    \begin{algorithmic}
        \item \textbf{Input:} point $\bm{p}$, robot SDF model $f_s$
        \item \textbf{Output:} joint configuration $\bm{q}'$ that satisfies $f_s(\bm{p}, \bm{q}')=\bm{0}$
        \item \textbf{Initialization:} $\bm{q}\leftarrow \bm{q_0}$ \Comment{Batch initialization}
        \item \textbf{for} $t = 1, \dots, T$ \Comment{T iterations}
        \item \hspace{2mm} $c \leftarrow c(\bm{q})$ \Comment{Compute cost}
        \item \hspace{2mm} $ \delta \bm{q} \leftarrow -\bm{H}^{-1} \nabla_{\bm{q}} c$ \Comment{Batch L-BFGS update}
        \item \hspace{2mm} $\bm{q} \leftarrow \bm{q} + \bm{\alpha} \delta \bm{q}$ \Comment{Line search}
        \item \textbf{end}
        \item $\bm{q}' \leftarrow \bm{q}: f_s(\bm{p}, \bm{q})< \epsilon$ \Comment{Return final configurations}

    \end{algorithmic} \vspace{-1mm}
\end{algorithm}

In this section, we introduce CDF and delve into its properties. We then present an efficient algorithm to compute CDF, as well as a fusion strategy for online combination of multiple CDFs.

\subsection{Problem Formulation}

CDF is inspired by recent work encoding SDF with robot joint configurations~\cite{liu2022regularized,koptev2022neural,li2024representing}. Let $r(\bm{q}) $ denote a robot at configuration $\bm{q} \in \mathbb{R}^n$ and $\bm{p} \in \mathbb{R}^3 $ be a point set in the robot workspace, for a robot with $n$ degrees of freedom (DoF). The robot SDF $f_s$ is a function of $\bm{p}$ and $\bm{q}$ that measures the distance from $\bm{p}$ to the closest point on the robot surface $\partial r(\bm{q})$\footnote{All variables support batch operations, i.e. $\bm{p} \in \mathbb{R}^{b_1 \times 3}$ and $\bm{q} \in \mathbb{R}^{b_2 \times n}$ accounting for $\bm{f}_s \in \mathbb{R}^{b_1 \times b_2}$.}:
\begin{equation}
    \label{eq:sdf}
    f_s(\bm{p}, \bm{q}) = \pm\min_{\bm{p}' \in \partial r(\bm{q})} \|\bm{p} - \bm{p}'\|,
\end{equation}
where $\pm$ indicates the sign of the distance, which is positive if $\bm{p}$ is outside, zero on the surface, and negative otherwise. The differentiability of the robot SDF with respect to both $\bm{p}$ and $\bm{q}$ enables various gradient-based manipulation planning tasks.

The robot SDF representation encodes the robot geometry through forward kinematics, where the distance is Euclidean in the workspace but highly nonlinear in configuration space. In contrast to using task space distances, CDF is defined as a function $f_c$ that measures the minimal distance in radians from $\bm{q}$ to zero-level-set joint configurations $\bm{q}':f_s(\bm{p}, \bm{q}')=\bm{0}$  at $\bm{p}$, which would establish contact between the robot and the point:
\begin{equation}
    \label{eq:CDF}
    f_c(\bm{p}, \bm{q}) = \min_{\bm{q}'} \|\bm{q} - \bm{q}'\|.
\end{equation}

This distance in radians corresponds to the movement of joint angles, where the constraint $f_s(\bm{p},\bm{q}')\!=\!0$ implicitly solves the inverse kinematics problem by finding the configuration set $\bm{q}'$ on the zero-level-set of robot SDF model, given a point $\bm{p}$. CDF is unsigned according to this definition, as we focus more on the value of distance and gradient, where the sign can be determined either by combining it with SDF or by estimating the normal direction on boundary samples. The derivative of CDF with respect to $\bm{q}$ corresponds to joint velocity.

\subsection{Properties of CDF}
An SDF satisfies the eikonal equation $\| \nabla_{\bm{p}} f_s(\bm{p}, \bm{q})\|=1$ almost everywhere. Thus, the closest point on the robot surface to $\bm{p}$ can be calculated by projecting $\bm{p}$ along the gradient direction: 
\begin{equation}
    \label{eq:sdf_projection}
    \bm{p}' = \bm{p} - f_s(\bm{p}, \bm{q}) \; \nabla_{\bm{p}} f_s(\bm{p}, \bm{q}).
\end{equation}

For surface points, gradients correspond to normal directions. Similarly, CDF satisfies the eikonal equation $\| \nabla_{\bm{q}} f_c(\bm{p}, \bm{q})\|=1$ almost everywhere in configuration space. The closest configuration on the zero-level-set manifold can be found by projecting the current configuration along the gradient direction:
\begin{equation}
    \label{eq:cdf_projection}
    \bm{q}' = \bm{q} - f_c(\bm{p}, \bm{q}) \; \nabla_{\bm{q}}f_c(\bm{p}, \bm{q}).
\end{equation}

This property makes CDF useful in manipulation planning tasks. It allows for direct computation of zero-level-set joint configurations through gradient projection, efficiently solving the inverse kinematics problem in one-step computation. For motion generation tasks, this implies having a more structured distance field in the configuration space, where gradients always point toward objects to reach or away from obstacles. Moreover, geodesics in the configuration space will naturally curve around the zero-level-sets, which can for example be used to move around an object while maintaining a constant joint angle distance to the object. From a control perspective, it means that the object remains reachable/avoidable within the robot joint angle velocity limits.

\subsection{Computation of CDF}\label{sec:CDF_computation}
The derivation of CDF is based on Eq.~\eqref{eq:CDF}, involving three components: 1. constructing the SDF model $f_s$ of the robot; 2. given a point $\bm{p}$, calculating zero-level-set configurations $\bm{q}'$ that satisfy $f_s(\bm{p}, \bm{q}')=0$; 3. \textcolor{black}{Given current joint configuration $\bm{q}$,} finding the closest configuration \textcolor{black}{on the zero-level-set $\bm{q}'$}, coupled with the calculation of the $\ell^2$ norm distance to yield the CDF value. We will discuss each step in detail.

\subsubsection{Robot SDF model} 
Various approaches exist for calculating the signed distance from a point in the robot workspace to the robot surface. Early approaches involve representing the robot geometry using spheres or meshes to approximate a coarse SDF. Recent investigations employ deep neural networks~\cite{liu2022regularized,koptev2022neural} for encoding the robot SDF. We adopt the method presented in~\cite{li2024representing} that exploits kinematic chains and basis functions to represent the robot SDF $f_s$, trading off accuracy and efficiency by providing a balance between explicit and implicit representation. 

\subsubsection{Finding zero-level-set configurations}\label{sec:zero-level-set}
The challenge of determining zero-level-set configurations parallels the inverse kinematics (IK) problem. While IK only focuses on the end-effector, CDF provides a more expressive approach that focuses on the whole robot geometry. We cast it as an optimization problem and employ the L-BFGS algorithm~\cite{nocedal1999numerical}. L-BFGS is a quasi-Newton method that has demonstrated effectiveness in robot motion planning~\cite{sundaralingam2023curobo}. The choice of L-BFGS is motivated by its relative simplicity and efficient parallelization. Alternative methods based on Gauss-Newton optimization could also be chosen. We formulate the cost function as a squared sum of SDF values, denoted as $c = \sum f_s^2(\bm{p},\bm{q})$. The search direction is updated using standard L-BFGS steps, and a line search approach is conducted for stable updates. The algorithm, outlined in Algorithm~\ref{algorithm1}, involves initializing a batch of joint configurations $\bm{q}$, by concurrently optimizing them to establish dense zero-level-set configurations. Additionally, our SDF model provides the link index of the robot in contact with the point $\bm{p}$ at configuration $\bm{q}'$, offering valuable information for subsequent computations.

\subsubsection{Retrieving CDF value}
Given an input point $\bm{p}$ and configuration $\bm{q}$, the procedure outlined in {\color{black} Section}~\ref{sec:CDF_computation} {\color{black} step 2} identifies zero-level-set configurations $\bm{q}'$ corresponding to $\bm{p}$. Calculating the CDF value involves determining the closest $\bm{q}'$ from $\bm{q}$. However, the sparse sampling of $\bm{q}'$ may result in an overly smooth CDF. To mitigate this, we reformulate \eqref{eq:CDF} as

\begin{equation}
    \label{eq:CDF_modified}
    f_c(\bm{p}, \bm{q}) = \min_{k=1,\ldots,K} (\min_{ \bm{q}'}\|\bm{q}_{:k} - \bm{q}'_{:k}\|),
\end{equation}
where $k$ denotes the $k$th robot link in contact with $\bm{p}$, $K$ is the total number of robot links, and $\bm{q}_{:k}$ represents all joint configurations before link $k$. This adjustment is rooted in the observation that CDF is influenced solely by preceding joint angles before the contact link. This modification exploits the inherent kinematic structure of the robot, leading to a more accurate approximation of CDF and reducing uncertainty, especially when $\bm{q}'$ samples are limited. The corresponding gradient of CDF is expressed as
\begin{equation}
    \label{eq:CDF_gradient}
    \centering
    \begin{aligned}
    \bm{q'}_{\min},k_{c} = \underset{\bm{q}',k}{\arg \min } \|\bm{q}_{:k} - \bm{q}'_{:k}\|,
    \\
    \nabla_{\bm{q}} f_c(\bm{p}, \bm{q}) =\frac{ \bm{q}_{:k} - \bm{q}'_{\min,:k_{c}}}{\|\bm{q}_{:k} -\bm{q}'_{\min,:k_{c}}\|},
    \end{aligned}
\end{equation}
where $\bm{q'}_{\min}$ is the closest joint configuration and $k_{c}$ is the corresponding contact link. The gradient possesses a unit $\ell^2$ norm and points against the direction of the nearest joint configuration on the zero-level-set.

\subsection{Fusion of CDF}
The computation of CDF described in Section~\ref{sec:CDF_computation} is applicable to both single points and batches of points. However, this process typically requires 1--10 seconds to find joint configurations and is scene-dependent. To address this challenge, we introduce a fusion strategy that computes the CDF independently for each point and combines them, yielding a scene-agnostic CDF representation conducive to efficient online calculations. Specifically, the point cloud $\bm{p}$ with $N$ points can be partitioned into $M$ subsets ($M \leq N$):
\begin{equation}
    \label{eq:CDF_fusion_1}
    \bm{p} = \{\bm{p}^1, \cdots, \bm{p}^M\},
\end{equation}
where $\bm{p}^i$ represents a subset of $\bm{p}$ with $N_i$ points. The CDF $f_c$ is constructed by fusing the CDFs $f_c^i$ of each subset:
\begin{equation}
    \label{eq:CDF_fusion_2}
    f_c(\bm{p}, \bm{q}) = \min_{i=1,\ldots,M} f_c^i(\bm{p}^i, \bm{q}).
\end{equation}

\begin{figure}
    \centering
    \includegraphics[width=1.0\linewidth]{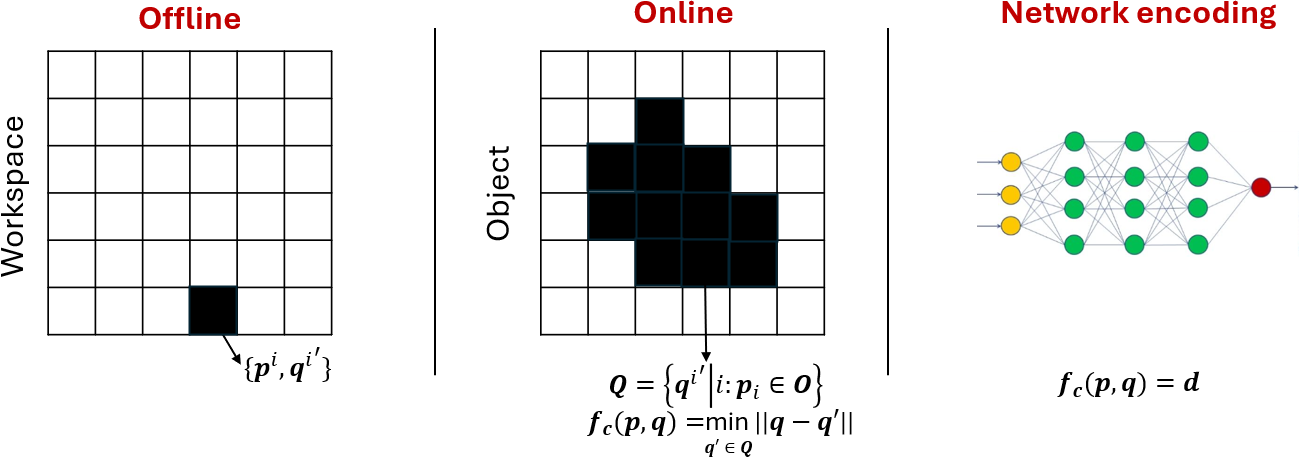}
    \caption{\textcolor{black}{Illustration of the computation of CDF. During the offline phase, we initialize the workspace of the robot as a volumetric grid and compute zero-level-set joint configurations for each grid point. For online computation, given an object $\mathcal{O}$, we identify the closest configuration in the set $\bm Q$ associated with occupied grids to calculate the $\ell^2$ distance. We further encode the CDF with neural networks to obtain a compact and grid-free representation.}}
    \label{fig:computation}
\end{figure}

For the extreme case where $M=N$, each subset contains only one point, allowing for offline computation and storage. The online inference stage only involves subtraction and minimum operations to fuse the CDFs based on the input, ensuring simplicity and efficiency. For example, initializing the workspace into a Cartesian grid, pre-computing corresponding joint configurations for each grid cell, and updating occupied cells during scene changes (see Figure~\ref{fig:computation}). This fusion strategy enables a non-parametric CDF representation and can be generalized to arbitrary environments.

The fusion of CDF also connects to the union operation of SDFs,
albeit performed in configuration space. Consequently, other Boolean operations used to compose and transform SDFs can also be applied to CDF, such as subtraction, intersection, repetition, and rounding. 

\section{Neural Configuration Space Distance Field}
\label{sec:lCDF}

In this section, we elaborate on the extension of the CDF through a learning-based approach to formulate an implicit representation, referred to as neural CDF. In contrast to the online computation detailed in Section~\ref{sec:CDF}, employing neural networks for CDF offers additional advantages. It disentangles from spatial resolution constraints, allowing for an expressive representation with reduced memory requirements. The neural CDF, being grid-free, facilitates distance queries between arbitrary joint configurations and points, enhancing flexibility and efficiency. Additionally, it presents a continuous representation, providing access to analytical gradients. Lastly, neural CDF operates in latent space and serves as a feature extractor for downstream tasks. In summary, neural CDF introduces trade-offs between accuracy, efficiency, and compression capabilities while enhancing flexibility.

Neural CDF approximates the batched function $\bm{f}_c(\bm{p}, \bm{q}): \mathbb{R}^{b_1 \times 3} \times \mathbb{R}^{b_2 \times n}\rightarrow \mathbb{R}^{b_1 \times b_2}$ by learning the weights of a multilayer perceptron (MLP) network. It takes the concatenation of $\bm{p}$ and $\bm{q}$ as input, with size $\mathbb{R}^{b_1b_2 \times (3+n)}$ and outputs the CDF value $\mathbb{R}^{b_1b_2\times 1}$. The neural CDF remains scene-agnostic, allowing the straightforward online fusion of different points. Subsequent sections will delve into data generation procedures, loss function design, training, and learning results.

\subsection{Dataset Generation}
The dataset generation process aligns with the computational and fusion procedures detailed in Section~\ref{sec:CDF}, comprising both offline and online components. In the offline phase, we construct a \textcolor{black}{$T\!\times\!T\!\times\!T$} volumetric grid in the 3D robot workspace. Utilizing Algorithm~\ref{algorithm1}, joint configurations $\bm{q}'$ that satisfy $f_s(\bm{p}, \bm{q}')=0$ for each grid point $\bm{p}$ are computed. Subsequently, a farthest point sampling algorithm is applied to downsample the obtained configurations. The resulting zero-level-set configurations for each grid point serve as templates for online computations. In the online phase, $b_1$ points and $b_2$ joint configurations, randomly sampled within joint limits, are selected. The closest template is identified, and the $\ell^2$ norm distance is computed using \eqref{eq:CDF_modified}. Simultaneously, the gradient concerning the joint configuration is calculated. The dataset generation process is outlined in Algorithm~\ref{algorithm2}.

\begin{algorithm}[t]
    \caption{Neural CDF Data Generation}
    \label{algorithm2}
    \begin{algorithmic}
        \item \textbf{Initialization:} volumetric grid $G$
        \item \text{\#\#\#  offline data}
        \item \textbf{for} each $p \in G$: \Comment{For each point on the grid}
        \item \hspace{2mm} $\bm{q}' \leftarrow \bm{q}: f_s(\bm{p}, \bm{q})=\bm{0}$ \Comment{Find $\bm{q}'$ using Algorithm~\ref{algorithm1}}
        \item \hspace{2mm} $\bm{q}' \leftarrow \text{Downsample}(\bm{q'})$ \Comment{Downsample $\bm{q}$}
        \item \text{\#\#\# online data}
        \item \textbf{for} $t = 1, \dots, T$ \Comment{Iterate over T epochs}
        \item \hspace{2mm} $\bm{p}, \bm{q}' \leftarrow \text{SampleOffline}()$ \Comment{Sample $\bm{p},\!\bm{q}'$ from offline data}
        \item \hspace{2mm} $\bm{q} \leftarrow \text{RandomSample}()$ \Comment{Online sample $\bm{q}$ that satisfies joint limits}
        \item \hspace{2mm} Compute $f_c, \nabla_{\bm{q}}f_c$  using~(\ref{eq:CDF_modified}) and~(\ref{eq:CDF_gradient}) \Comment{Ground truth}
        \item \hspace{2mm} $\cdots$
        \item \hspace{2mm} \text{ComputeLoss}()  \Comment{Network training}
        \item \hspace{2mm} $\cdots$
        \item \textbf{end for}
    \end{algorithmic} \vspace{-1mm}
\end{algorithm}

\subsection{Loss Function}
We design a loss function for training the neural CDF based on existing neural SDF representations~\cite{park2019deepsdf,gropp2020implicit,ortiz2022isdf}. The loss function consists of four components: distance loss, gradient loss, eikonal loss and tension loss, each serving a distinct purpose.

\textbf{Distance loss.} The distance loss is characterized by the mean squared error between the predicted CDF and the ground truth, expressed as
\begin{equation}
    \label{eq:loss_distance}
    \mathcal{L}_{\text{dist}} = \frac{1}{b_1 b_2} \sum_{i=1}^{b_1} \sum_{j=1}^{b_2} \big(\hat{f_c}({p}_i, {q}_j) - f_c({p}_i, {q}_j)\big)^2,
\end{equation}
where $\hat{f_c}$ and $f_c$ denote the predicted and ground truth C-space distances for point $p_i$ and configuration $q_j$, respectively.

\textbf{Gradient loss.} This term constrains the gradient of the predicted CDF to consistently point against the direction of the closest joint configuration on the zero-level set. It employs cosine similarity loss to penalize deviations, given by
\begin{equation}
    \label{eq:loss_gradient}
    \mathcal{L}_{\text{grad}} = \frac{1}{b_1 b_2} \sum_{i=1}^{b_1} \sum_{j=1}^{b_2} \left(1 - \frac{{\nabla_{\bm{q}}\hat{f_c}({p}_i, {q}_j)}^{\!\trsp} \;\nabla_{\bm{q}}f_c({p}_i, {q}_j)}{\|\nabla_{\bm{q}}\hat{f_c}({p}_i, {q}_j)\| \; \|\nabla_{\bm{q}}f_c({p}_i, {q}_j)\|}\right).
\end{equation}

\textbf{Eikonal loss.} This term regulates the predicted CDF by encouraging its gradients to have a unit $\ell^2$ norm. This regularization, inspired by the eikonal partial differential equation, ensures a valid signed distance field~\cite{gropp2020implicit,ortiz2022isdf}. The eikonal regularization term is formulated as
\begin{equation}
    \label{eq:loss_eikonal}
    \mathcal{L}_{\text{eikonal}} = \frac{1}{b_1 b_2} \sum_{i=1}^{b_1} \sum_{j=1}^{b_2} \Big| \|\nabla_{\bm{q}}\hat{f_c}({p}_i, {q}_j)\| - 1 \Big|.
\end{equation}

\textbf{Tension loss.} The tension loss term aims to regularize the curvature of the CDF, promoting smoothness. It penalizes the squared sum of the Laplacian, which measures the second derivatives of the predicted CDF~\cite{juttler2002least,yariv2021volume}, namely
\begin{equation}
    \label{eq:loss_tension}
    \mathcal{L}_{\text{tension}} = \frac{1}{b_1 b_2} \sum_{i=1}^{b_1} \sum_{j=1}^{b_2} \| \nabla_{\bm{q}}^2\hat{f_c}({p}_i, {q}_j)\|^2,
\end{equation}
where $\nabla_{\bm{q}}^2$ is the Laplacian operator computed via automatic differentiation.

\textbf{Total loss.} The network is optimized to minimize the weighted sum of the four loss terms
\begin{equation}
    \label{eq:loss_total}
    \mathcal{L}_{\text{total}} = \lambda_1 \mathcal{L}_{\text{dist}} + \lambda_2 \mathcal{L}_{\text{grad}} + \lambda_3 \mathcal{L}_{\text{eikonal}} + \lambda_4 \mathcal{L}_{\text{tension}},
\end{equation}

In the experiments, we set $\lambda_1\!=\!5.0, \lambda_2\!=\!0.1, \lambda_3\!=\!0.01, \lambda_4\!=\!0.01$. 

\subsection{Implementation Details}
For the training of the neural CDF model, we employ a simple fully connected MLP. To assess its effectiveness in handling high-dimensional inputs, we evaluate the neural CDF on a 7-axis Franka robot. \textcolor{black}{The resolution of the volumetric grid $T$ is set to 20 for data generation.} The input dimension is $3\!+\!7$, and the output corresponds to the configuration space distance. In line with previous work~\cite{koptev2022neural}, we adopt a 5-layer MLP architecture, where the input data is enriched with position encoding~\cite{mildenhall2021nerf}. 
During training, we randomly sample $b_1=4000$ points with corresponding joint configurations and $b_2=100$ configurations. Thus, the batch size is $4000 \times 100$. The network is trained for $50,000$ epochs using the Adam optimizer with a learning rate of $0.001$, decayed by a factor of $0.5$. The training process spans approximately $2$ hours on a single NVIDIA RTX 3090 GPU. 

\begin{table*}[t]
    \begin{center}
        \begin{tabular}{ccc|c}
            \centering
            \includegraphics[width =0.23\linewidth
            ]{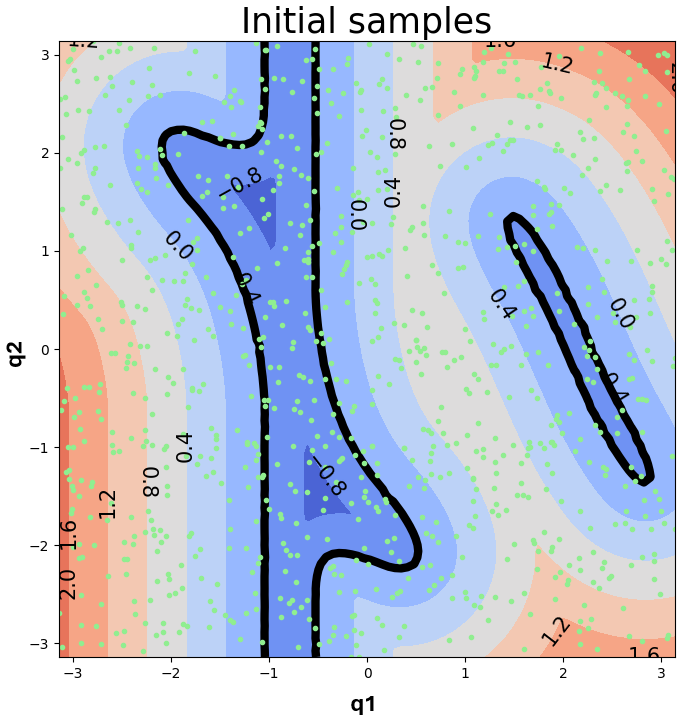}   &
            \includegraphics[width =0.23\linewidth]{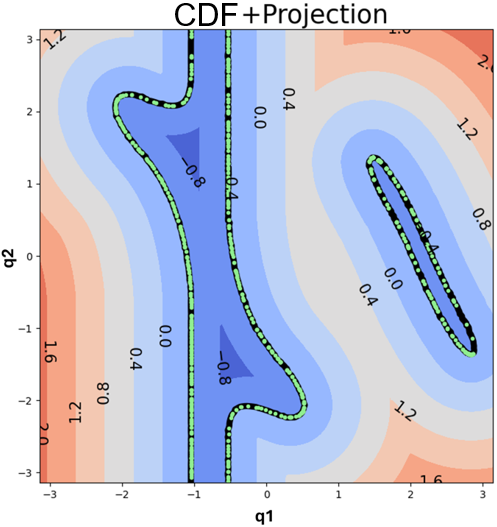}  &

            \includegraphics[width =0.20\linewidth]{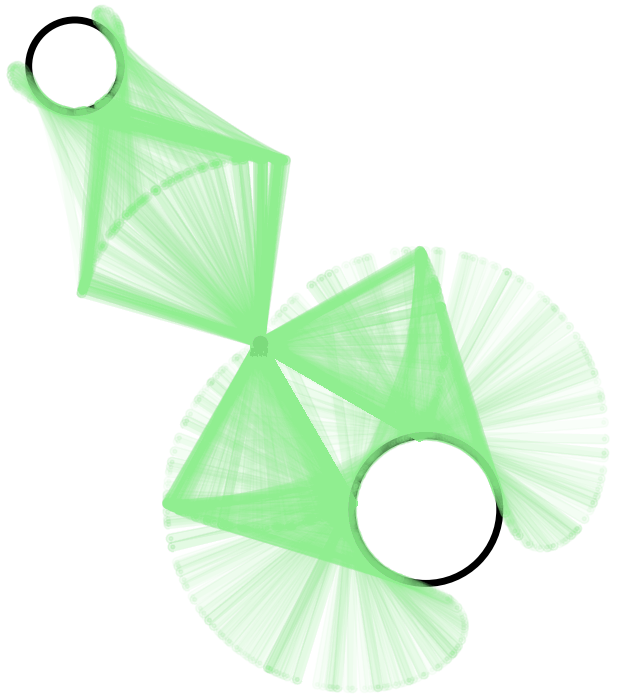}  &
            \includegraphics[width =0.23\linewidth]{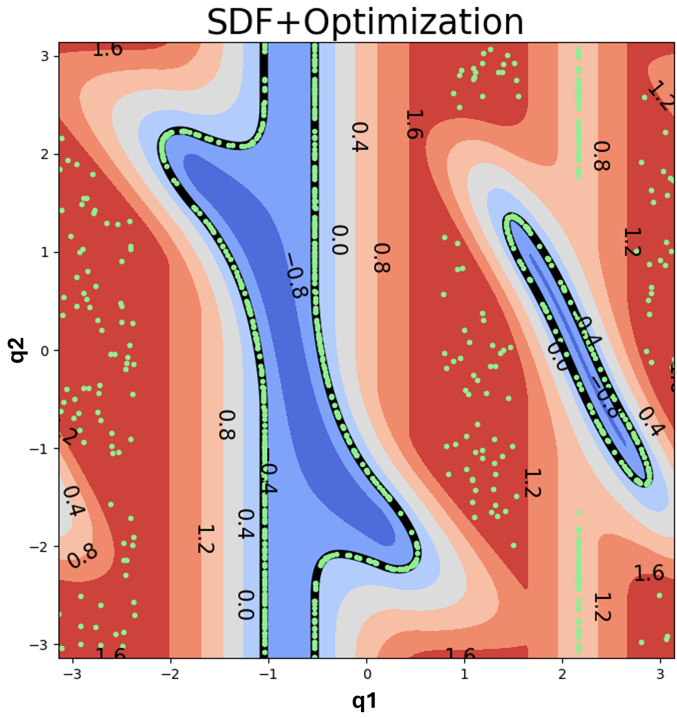}  \\
            (a) & (b) & (c) & (d)
        \end{tabular}
        \captionof{figure}{Comparison between CDF and SDF in solving whole-body inverse kinematics problem. (a) The initial sampled joint configurations. (b) Gradient projection by CDF. (c) Task space visualization of feasible solutions in (b). (d) Results for distance query-based method with L-BFGS optimizer. We can see that with the baseline SDF approach, the system can get stuck when the gradient of the SDF vanishes or reaches the singularity.
        }
        \label{fig:2d_proj}
    \end{center}
\end{table*}
\subsection{Learning Results}
We evaluate the trained neural CDF model through a comprehensive evaluation for both accuracy and efficiency. The results are presented in Table~\ref{tab:lCDF}. Specifically, we run the forward pass of the network, which outputs predicted C-space distance values $f_c$ for input pairs $\bm{p}$ and $\bm{q}$. Then we compute the gradient via automatic differentiation and project configurations $\bm{q}$ to points $\bm{p}$ along gradient direction using \eqref{eq:cdf_projection}. According to the definition of CDF, the distance between the robot surface, defined by projected configurations $\bm{q}_{\text{proj}}$ to input points $\bm{q}$, should be $0$. Thus, we measure the mean absolute error (MAE) and root mean squared error (RMSE) as metrics. The success rate (SR) denotes the percentage of configurations successfully projected to input points within a threshold of $3cm$. The projection process is designed to run iteratively for improved accuracy. Each experiment involves the random sampling of $1000$ points and $1000$ configurations (the results are reported as averages).
The outcomes reveal that our model accurately predicts CDF values and gradients, facilitating a projection process that successfully identifies the closest joint configurations on the zero-level-set. Stability is achieved after 2 iterations. As for computation time, results are provided for a single NVIDIA RTX 3090 GPU and a 30-core 2.2GHz CPU. Inference time denotes the duration for a single forward pass of the network, while projection time encompasses the time for automatic differentiation and the projection process. These results underscore the efficiency, high parallelizability, and scalability of our neural CDF, particularly when dealing with large batch sizes.

\begin{table}[]
    \centering
    \caption{Accuracy and computation time (GPU / CPU) of Neural CDF on the Franka robot.}
    \label{tab:lCDF}
    \begin{adjustbox}{width=\columnwidth,center}
    \begin{tabular}{cccc|ccc}
    \toprule
    \multicolumn{4}{c|}{Accuracy} & \multicolumn{3}{c}{Computation Time} \\
    \cline{1-4} \cline{5-7}
    Projection & \multirow{2}{*}{MAE (cm)}   & \multirow{2}{*}{RMSE (cm)}  & \multirow{2}{*}{SR(\%) }  & Batch & Inference & Projection \\
    Iterations & & & & Size  & Time (ms) & Time (ms) \\
    \hline
    1          & $4.99\textcolor{black}{\pm1.93}$ & $8.59\textcolor{black}{\pm3.15}$ & $60.3\textcolor{black}{\pm 12.20}$ & 1 & $0.49 / 0.37$ & $0.71 / 0.34$ \\
    2          & $1.64\textcolor{black}{\pm0.62}$ & $2.80\textcolor{black}{\pm1.20}$ & $87.8\textcolor{black}{\pm 9.50}$ & $10$ & $0.51 / 0.59$ & $0.72 / 0.66$ \\
    3          & $1.39\textcolor{black}{\pm0.51}$ & $2.09\textcolor{black}{\pm0.94}$ & $91.1\textcolor{black}{\pm 8.12}$ & $10^2$ & $0.56 / 0.95$ & $0.75 / 1.03$ \\
    4          & $1.36\textcolor{black}{\pm0.49}$ & $2.00\textcolor{black}{\pm0.89}$ & $91.6\textcolor{black}{\pm 8.23}$ & $10^3$ & $0.58 / 10.20$ & $0.97 / 5.48$ \\
    5          & $1.34\textcolor{black}{\pm0.48}$ & $1.92\textcolor{black}{\pm0.79}$ & $91.8\textcolor{black}{\pm 8.31}$ & $10^4$ & $0.79 / 25.00$ & $1.01 / 36.00$ \\
    10         & $1.35\textcolor{black}{\pm0.52}$ & $1.89\textcolor{black}{\pm0.80}$ & $91.6\textcolor{black}{\pm 8.39}$ & $10^5$ & $4.61 / 329.00$ & $11.30 / 310.00$ \\
    \bottomrule
    \end{tabular}
    \end{adjustbox}
\end{table}

\section{CDF for Whole-body Inverse Kinematics}
\label{sec:wik}

CDF inherently encodes the kinematic structure of the robot, offering a solution to the inverse kinematics problem through gradient projection without the need for iterative procedures. Given its holistic modeling of the robot geometry, our approach extends the inverse kinematics problem to whole-body inverse kinematics problems, instead of only focusing on the end-effector. We assess our approach with a planar robot and the 7-axis Franka robot.

\subsection{2-DoF Planar Robot}
We start with a straightforward example involving a 2D planar robot with link lengths $l_1\!=\!l_2\!=\!2$ and joint limits $q_1, q_2 \in [-\pi, \pi]$. The CDF is computed online using the methodology outlined in Section~\ref{sec:CDF}. Two circular objects with radii $r_1\!=\!0.8$ and $r_2\!=\!0.5$ are positioned at $(1.8, -1.8)$ and $(-2.0, 3.0)$, respectively. The objective of the whole-body inverse kinematics task is to identify joint configurations that make the robot reach the objects. We compare our CDF representation with SDF and present qualitative results in Figure~\ref{fig:2d_proj}. The results demonstrate that in this 2D scenario, CDF effectively solves the problem through a one-step gradient projection, while SDF-based optimization struggles to find solutions when the gradient vanishes and gets stuck in local minima due to the nonlinearity of the forward kinematics function.


\subsection{7-DoF Franka Robot}

\begin{figure*}[htbp]
    \centering
    \includegraphics[width = 1.0\linewidth]{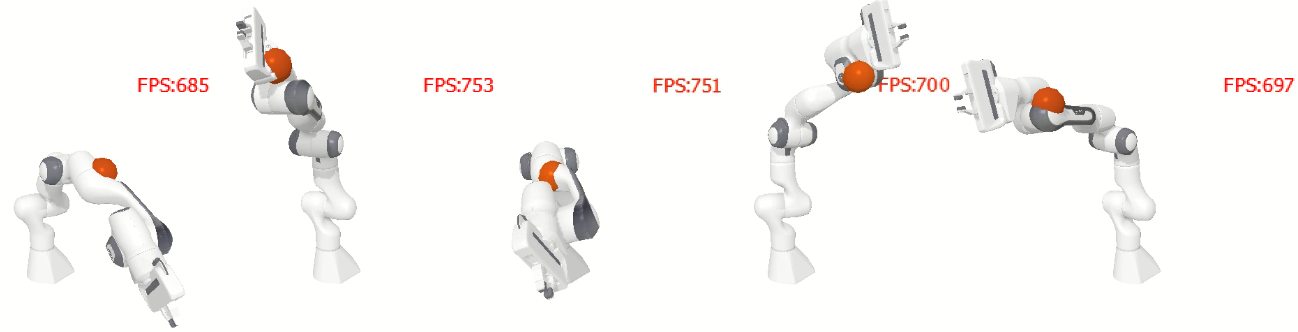}
    \caption{Gradient projection for whole-body inverse kinematics using neural CDF. The centers of the red spheres are target points, where the radius of spheres is set to $0.05m$.}
    \label{fig:lCDF}
\end{figure*}

To further evaluate the performance of CDF, we conducted experiments with a 7-axis Franka robot, utilizing the trained neural CDF model outlined in Section~\ref{sec:lCDF}. The evaluation focused on the whole-body inverse kinematics performance for various target points, employing $10'000$ randomly initialized configurations. The gradient projection process was iteratively performed in three steps to enhance performance. For comparison, two baseline representations were included in the evaluation: the whole-body SDF representation proposed in \cite{li2024representing} and the neural joint space SDF representation (Neural-JSDF) proposed in \cite{koptev2022neural}. Both approaches are followed with an L-BFGS algorithm for optimization, which is also described in cuRobo~\cite{sundaralingam2023curobo}, achieving state-of-the-art performance. 
Experiments are repeated $100$ times and average results are shown in Table~\ref{tab:ik_7d}. CDF demonstrated the ability to compute over $700'000$ valid solutions per second, outperforming the state-of-the-art distance query-based approach by $180$ times, which could only find $3'700$ solutions. Additionally, the inference and projection process of CDF took only $1\!-\!2$ milliseconds, with the primary time cost attributed to the post-processing of distance checking to select valid solutions that satisfy the error threshold. Figure \ref{fig:lCDF} shows the results of a 1-step projection for different target point positions.

\begin{table}[htbp]
    \centering
    \caption{Comparison of CDF and SDFs in whole-body inverse kinematics task with a 7-axis Franka robot. CDF solves $8773$ solutions in $10.6$ ms while the SDF based method only finds $3652$ solutions in $971$ ms.}
    \begin{tabular}{|c|c|c|c|}
    \hline
    \textbf{Methods} & \textbf{Valid Solutions} & \textbf{Time (ms)} \\
    \hline
    CDF + 1-step Projection & 6089 & \textbf{8.72} \\
    \hline
    CDF + 2-step Projection & 8773 & 10.60 \\
    \hline
    CDF + 3-step Projection & \textbf{9163} & 12.70 \\
    \hline
    SDF + L-BFGS optimizer & 3652 & 971.00 \\
    \hline
    Neural-JSDF + L-BFGS optimizer& 264 & 272.00 \\
    \hline
    \end{tabular}
    \label{tab:ik_7d}
\end{table}

\begin{figure*}[htbp]
    \centering
    \includegraphics[width=1.0\linewidth]{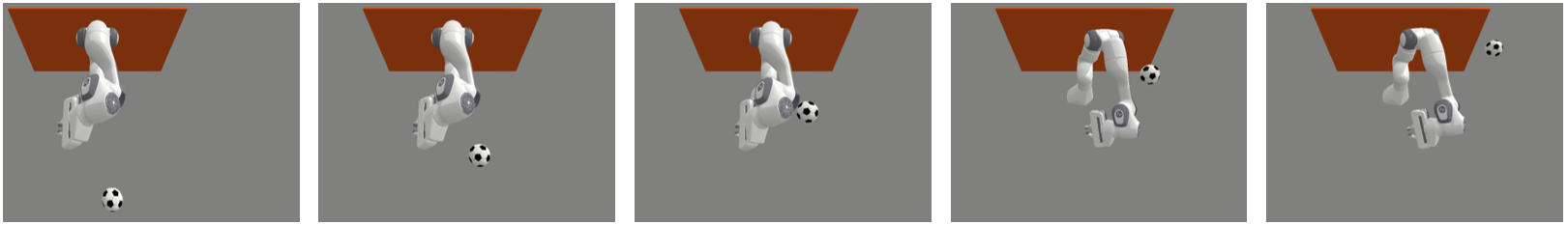}
    \caption{Goalkeeper task in simulation.}
    \label{fig:ball}
\end{figure*}

\begin{figure*}
    \centering
    \includegraphics[width=1.0\linewidth]{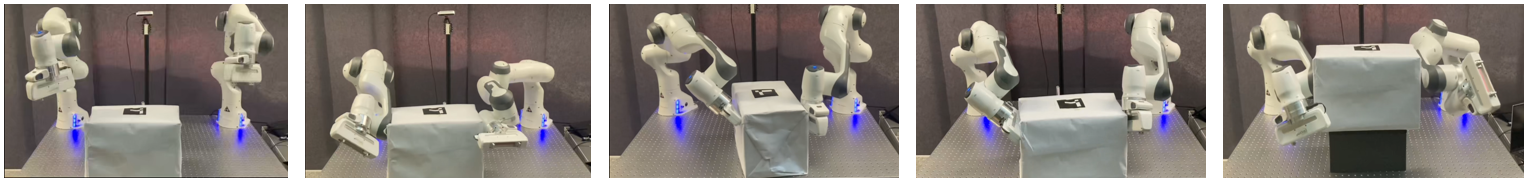}
    \caption{Planned configurations to reach the box. The first image shows the initial configurations of the two arms.}
    \label{fig:box}
\end{figure*}
\subsection{Applications}
We demonstrate two applications that leverage the gradient projection capabilities of CDF. The first application involves a goalkeeper task where the robot intercepts a thrown ball using its arm links. The second one is a dual-arm lifting task exploiting the whole-body structure of the robot to establish contact with a large box, which is hard to accomplish conventionally using an end-effector.

\subsubsection{Goalkeeper task}
In contrast to tasks involving rapid robot responses to avoid obstacles, the present task entails the robot acting as a goalkeeper, utilizing its arm to intercept a propelled ball. This task poses increased difficulty as the robot must promptly determine a whole-body inverse kinematics solution and transition to the requisite configuration to intercept the ball. The experimental configuration is outlined as follows: (1) a rectangular goal, measuring $0.8m$ in width and $0.6m$ in height, is positioned behind the robot, whose configuration is initialized in the middle of the joint angle range; (2) a ball is thrown toward the goal from the front of the robot with a randomly assigned direction and velocity; (3) the robot is tasked to move its arm to intercept the ball.

The conducted evaluations are performed in a simulated environment, with the assumption that the robot can only perceive the current position of the ball, necessitating swift movements to ensure an effective defense. A joint position controller is employed to govern the robot arm, directing it to the designated joint configuration. The task is executed 100 times, resulting in an 82\% success rate for our CDF representation. In comparison, the SDF-based method achieves a success rate of only 35\%. Snapshots of our approach are depicted in Fig.~\ref{fig:ball}. 

\subsubsection{Large box lifting} 

The objective of this task is to plan joint configurations for two robot arms to establish contact with a designated box. We assume that the contact points on the box are predefined, and the robots can use any surface points on their body for establishing contact. This task typically involves a multi-objective optimization problem with constraints, including joint limits, collision avoidance, and goal-reaching. The combination of these objectives introduces non-convexity and makes the problem hard to solve. Leveraging the efficient and parallelizable gradient projection inherent to CDF, we instead present a straightforward sample-filter approach to address this problem. Specifically, we iteratively sample a batch of initial configurations, project them onto the contact points, and filter out configurations in collision with the box or violating joint limits. This process continues until feasible solutions satisfying all constraints are identified. An evaluation of our approach, compared with the Gauss-Newton optimization method outlined in~\cite{li2024representing}, is presented in Table~\ref{tab:whole_arm_lifting}. The results indicate that CDF reduces the planning time by a factor of $7$ and generates shorter paths. During the lifting phase, the Jacobian matrix of the contact point w.r.t.~the joint configuration is computed. A joint impedance controller is used in the experiment. Qualitative results are shown in Fig.~\ref{fig:box}.

\begin{table}[htbp]
    \centering
    \caption{Comparison results on large box lifting task.}
    \begin{tabular}{c|c|c}
    \hline
    \textbf{Methods} & \textbf{Planning Time(s)} & \textbf{Average Distance(rad)} \\
    \hline
    CDF + Filter & \textbf{7.65} & \textbf{1.37} \\
    \hline
    SDF + Optimizer & 54.80 & 2.85 \\
    \hline
    \end{tabular} 
    \label{tab:whole_arm_lifting}
\end{table}

\section{CDF for Manipulation Planning}
\label{sec:mp}

\begin{figure*}[htbp]
    \begin{center}
        \begin{tabular}{cc|cc}
            \centering
            \subfigure[CDF C-Space]{%
                \includegraphics[width =0.22\linewidth]{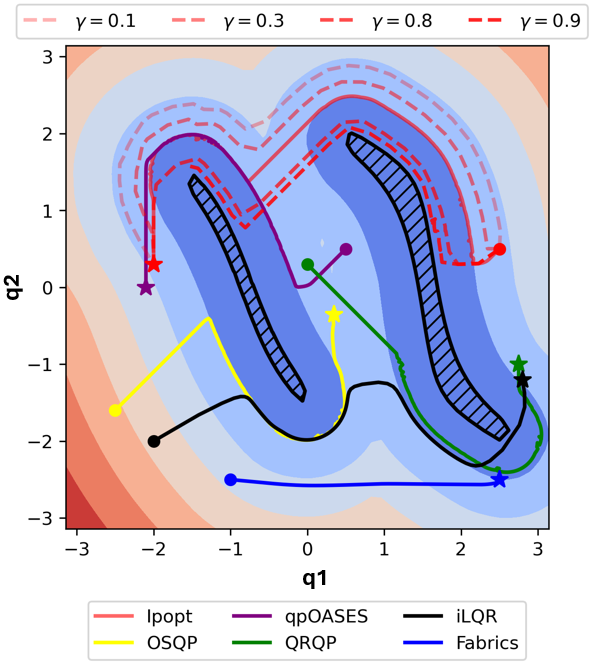}
                \label{fig:MP_cdf_c_space}
            }
            &
            \subfigure[SDF C-Space]{%
                \includegraphics[width =0.22\linewidth]{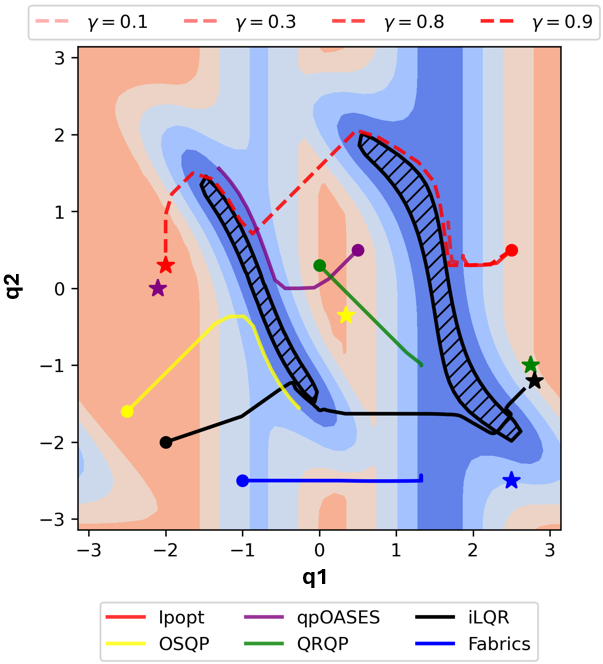}
                \label{fig:MP_sdf_c_space}
            }
            &
            \subfigure[CDF T-Space]{%
                \includegraphics[width =0.24\linewidth]{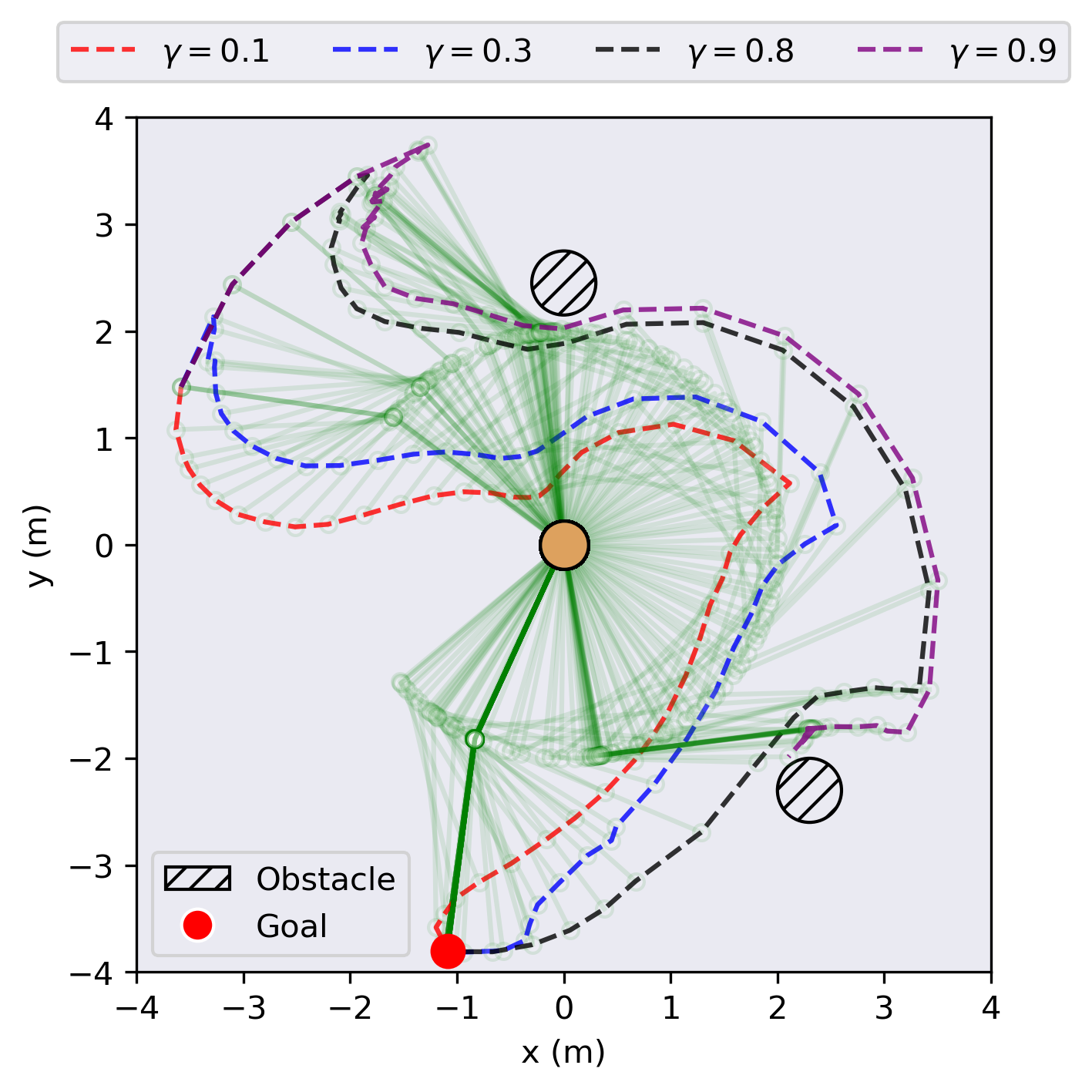}
                \label{fig:MP_cdf_t_space}
            }
            &
            \subfigure[SDF T-Space]{%
                \includegraphics[width =0.24\linewidth]{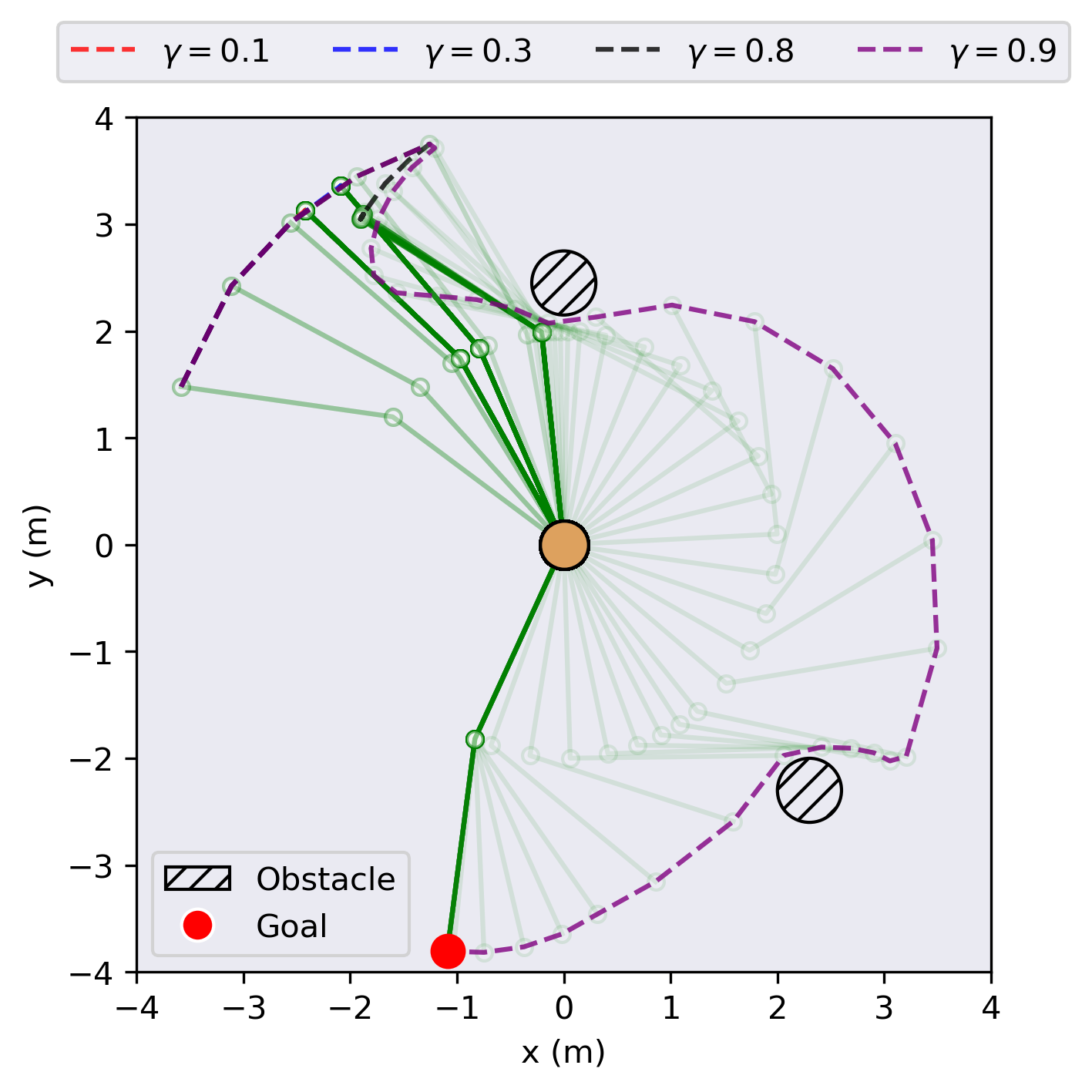}
                \label{fig:MP_sdf_t_space}
            }
        \end{tabular}
        \caption{CDF/SDF-based motion planning approaches. Different methods are shown in different colors. We also demonstrate how the safety buffer affects the planning results of CDF and SDF. 
        }
        \label{fig:MP_2d_demonstration}
    \end{center}
\end{figure*}

\begin{table*}[ht!]
\centering
\caption{Experiments for motion planning on SDF and CDF.}
\resizebox{\textwidth}{!}{
\begin{tabular}{c|ccc|ccc|ccc|ccc}
\hline
\multirow{2}{*}{} & \multicolumn{3}{c|}{2D - CDF}               & \multicolumn{3}{c|}{2D - SDF}                   & \multicolumn{3}{c|}{7D - CDF}               & \multicolumn{3}{c}{7D - SDF}                      \\ \cline{2-13} 
                  & Success Rate & Tracking Error (cm) & Time Step & Success Rate & Tracking Error (cm) & Time Step & Success Rate & Tracking Error (cm)& Time Step & Success Rate & Tracking Error (cm) & Time Step \\ \hline
IPOPT             & 92\%         & 1.27            & 253           & 51\%         & 2.19            & 284           & 93\%         & 0.98            & 231           & 51\%         & 1.12            & 290           \\
QRQP              & 94\%         & 1.16            & 245           & 52\%         & 2.07            & 279           & 91\%         & 0.99            & 226           & 38\%         & 1.31            & 297           \\
OSQP              & 88\%         & 1.21            & 300           & 40\%         & 2.19            & 321           & 91\%         & 0.92            & 279          & 48\%         & 1.04            & 301           \\
qpOASES           & 91\%         & 1.04            & 241           & 63\%         & 1.74            & 263           & 93\%         & 0.99            & 229           & 51\%         & 1.13            & 289           \\ \hline
Geometric Fabrics & 95\%         & 1.03            & -              & 68\%         & 1.76            & -              & 88\%         & 2.02            & -              & 74\%         & 2.18            & -              \\ \hline
iLQR              & 76\%         & 0.06            & -              & 55\%         & 0.12            & -              & 48\%         & 0.02            & -              & 38\%         & 0.03            & -              \\ \hline
\end{tabular}
}
\label{tab:MP_bench}
\end{table*}

In this section, we investigate the use of CDF for manipulation planning tasks. The key advantage of CDF is the structured representation that alleviates challenges arising from nonlinearity and singularity, making motion optimization in configuration space easier. Similarly to conventional SDF, CDF provides efficient queries of distances and gradients, enabling large-scale parallel computation. To demonstrate its efficacy, we initially explore qualitative results through 2-DOF examples and then progress to 7-DOF robot scenarios, including real-world experiments. 

\subsection{Benchmark Approaches and Evaluation metrics}
We evaluate the CDF representation on several gradient-based motion optimization approaches:

\subsubsection{Quadratic programming} We first formulate the motion planning task as reactive quadratic programming (QP) problem, drawing inspiration from the work of Mirrazavi et al.~\cite{mirrazavi2018unified}. The QP formulation is:
\begin{subequations}
\begin{align}
\boldsymbol{u}_k^{*} =\; &\underset{\boldsymbol{q},\boldsymbol{u}}{\arg\min} \;
 \boldsymbol{e}(\boldsymbol{q}_k)^{\top} \bm{H} \boldsymbol{e}(\boldsymbol{q}_k) + {\boldsymbol{u}_k}^{\top} \bm{R} \boldsymbol{u}_k, \label{eq: qp_objective_function}
\\
\text { s.t. } \;
& \boldsymbol{q}_{k+1}=\bm{A} \boldsymbol{q}_k+\bm{B} \boldsymbol{u}_k, \\
& \boldsymbol{q}_k \in \mathcal{Q},\; \boldsymbol{u}_k \in \mathcal{U}, \\
& -\nabla_{\boldsymbol{q}} f_c(\boldsymbol{p},\boldsymbol{q}) \boldsymbol{u}_k \Delta t \leq \ln (f_c(\boldsymbol{p}, \boldsymbol{q})+ \gamma), \label{eq: qp_collision_avoidance}
\end{align}
\end{subequations}
where $\boldsymbol{q}_k$ and $\boldsymbol{u}_k$ are the state and control input at time step $k$, $\bm{H}$ and $\bm{R}$ are the positive definite matrices for tracking errors and control efforts, $\boldsymbol{e}(\boldsymbol{q}_k) = \boldsymbol{q}_k - \boldsymbol{q}_{\text{desire}}$ is the error vector between the initial and goal configurations, $f_c(\boldsymbol{p},\boldsymbol{q})$ is the CDF, $\Delta t$ is the time step, $\gamma$ is a scalar hyperparameter that acts as a safety buffer, and $\mathcal{Q}$ and $\mathcal{U}$ are the admissible state and control constraints. The constraint~\eqref{eq: qp_collision_avoidance} ensures collision avoidance, where the robot is allowed to get close to the obstacle when far away, and it is forced to follow the tangent or normal direction of the gradient field when close. For implementation, we use the CasADi~\cite{Andersson2019} library and solve it with popular solvers including OSQP~\cite{stellato2020osqp}, qpOASES~\cite{ferreau2014qpoases}. Nonlinear programming solvers like IPOPT~\cite{wachter2006implementation} and QRQP~\cite{gillis2019nonlinear} are also tested.

\subsubsection{iterative Linear Quadratic Regulator (iLQR)} Another benchmark approach involves employing an iterative Linear Quadratic Regulator that solves the optimal control problem by iteratively linearizing the dynamics and cost function around the current trajectory~\cite{li2004iterative}. The dynamic system is defined as $\Delta \boldsymbol{q}_{k+1}=\bm{A} \Delta \boldsymbol{q}_k+\bm{B} \Delta \boldsymbol{u}_k$. We minimize the cost function
\begin{equation}
\begin{aligned}
c(\boldsymbol{q}, \boldsymbol{u}) &=  \boldsymbol{e}(\boldsymbol{q}_K)^{\top} \boldsymbol{Q}_1 \boldsymbol{e}(\boldsymbol{q}_K)  \\
&+ \sum_{k=1}^{K-1} \boldsymbol{h}(\boldsymbol{q}_k)^{\top} \boldsymbol{Q}_2\boldsymbol{h}(\boldsymbol{q}_k) + {\boldsymbol{u}_k}^{\top} \boldsymbol{R} \boldsymbol{u}_k 
\end{aligned}
\end{equation}
where $\boldsymbol{Q}_1$ and $\boldsymbol{Q}_2$ are 
precision matrices for tracking errors and collision avoidance, $\boldsymbol{R}$ is the control effort matrix.
$\boldsymbol{h}(\boldsymbol{q}_k) = \min(f_c(\boldsymbol{p},\boldsymbol{q})-\gamma,\;0)$ represents the collision avoidance term. The solution of iLQR can be computed either in batch or recursive form, see~\cite{RCFS} for details.

\subsubsection{Geometric fabrics} Geometric fabrics is a reactive acceleration-based control policy $\ddot{\boldsymbol{q}} = \pi (\boldsymbol{q}, \dot{\boldsymbol{q}})$.
$\ddot{\boldsymbol{q}}$ is computed through the motion of equation $\boldsymbol{M} \ddot{\boldsymbol{q}}+\boldsymbol{F}=0$, where $\boldsymbol{M}(\boldsymbol{q}, \dot{\boldsymbol{q}})$ and $\boldsymbol{F}(\boldsymbol{q}, \dot{\boldsymbol{q}})$ model the generalized mass matrix and external forces based on positions and velocities
, see~\cite{ratliff2020optimization} for details. The obstacle avoidance geometry is defined as $\boldsymbol{h} = \lambda \| \dot{\boldsymbol{q}} \|^{2} \nabla_{\boldsymbol{q}}\boldsymbol{\psi}(f_c(\boldsymbol{p}, \boldsymbol{q}))$.
When the robot gets close to the obstacle, the value $\boldsymbol{\psi} (\boldsymbol{p}, \boldsymbol{q})$ increases and repels the robot away from the collision boundary. 
The geometric fabrics are adapted from the open-source implementation of optimization fabrics~\cite{spahn2023dynamic}.

For a comprehensive evaluation, we estimate both CDF and SDF. The implementation of SDF is achieved by replacing the distance function $f_c$ with $f_s$. Several evaluation metrics are adopted for comparison.
\begin{itemize}
    \item {\textbf{Success Rate}}: the success rate shows the percentage of collision-free trajectories generated while reaching the goal. As the evaluation involves randomly sampling initial and goal configurations, for the cases when all algorithms failed, the corresponding samples were excluded when reporting the success rate. 

    \item {\textbf{Tracking Error}}: As reactive approaches may get stuck at a local minimum, we introduce the tracking error to measure the final $\ell^2$ norm distance between the final configuration and the desired configuration.
    
    \item {\textbf{Time Step}}: The time step denotes the average number of time steps the agent requires to reach the goal configuration.
\end{itemize}

\subsection{Planar Robot Test}

For the 2D experiment, we follow the previous section that defined a 2D planar robot with link lengths $l_1\!=\!l_2\!=\!2$ and joint limits $q_1, q_2 \in [-\pi, \pi]$. Two circle obstacles with radius of $0.3$ are placed at $(2.3, -2.3)$ and $(0.0, 2.45)$. We randomly sample initial and goal configurations for 100 cases and report the experimental results in Table~\ref{tab:MP_bench}.
It shows that the approach based on CDF has a higher success rate and better tracking accuracy than SDF. The average number of time steps is smaller, increasing efficiency. To further investigate the mechanism behind this, we visualize some cases in Fig.~\ref{fig:MP_2d_demonstration}-a,b. It shows that the structured distance field can benefit all approaches mentioned above, as their trajectories follow the geodesics of CDF. In contrast, the nonlinearity of the configuration space when using the conventional SDF makes the optimizer get stuck into local minima. Additionally, we visualize the planning results of IPOPT with different safety buffers $\gamma$ ranging from $0.1$ to $0.9$. The planner exhibits a more conservative behavior as the value of $\gamma$ decreases to ensure safety. With CDF, the planned trajectory scaled well with different $\gamma$. In contrast, for SDFs, the planner only finds a solution when $\gamma=0.9$ and the trajectory in joint space is very close to the obstacle. It is also reflected in the task space (Fig.~\ref{fig:MP_2d_demonstration}-c,d), where the robot reaches a singularity when it is close to the obstacle.

\begin{figure*}
    \centering
    \includegraphics[width=1.0\linewidth]{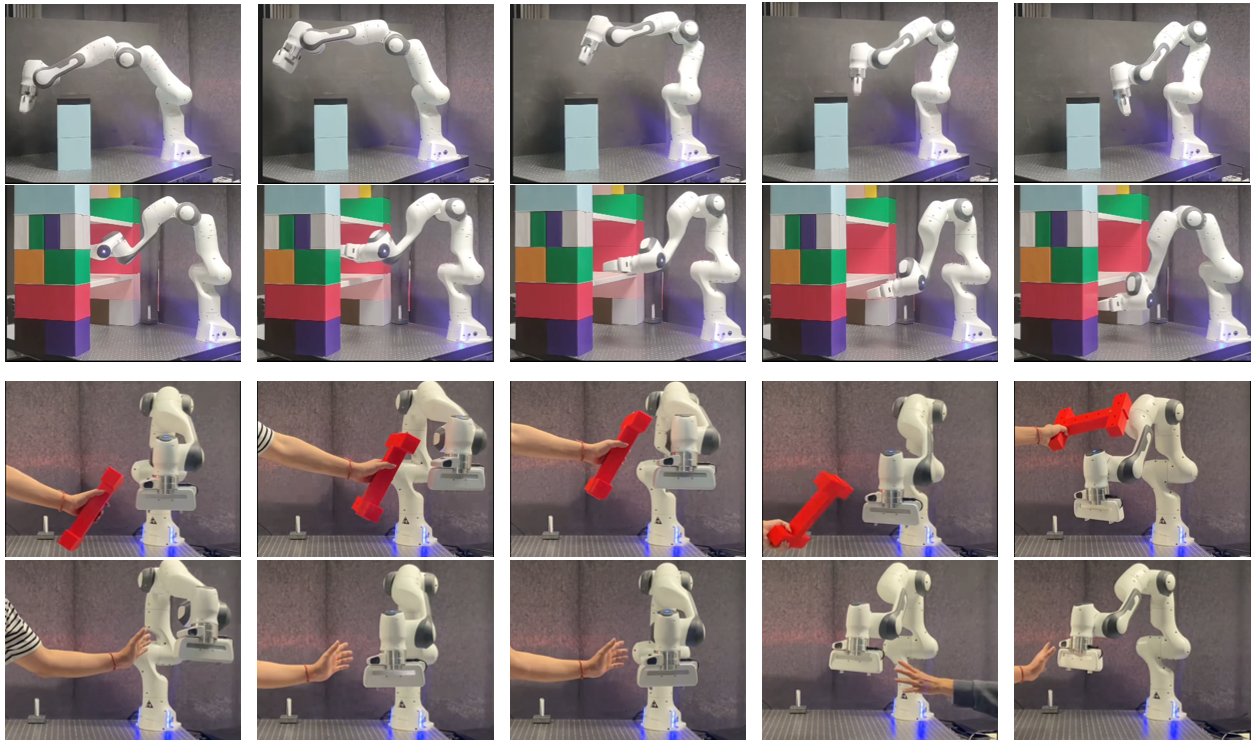}
    \caption{Motion planning and reactive collision avoidance using CDF. The top two rows show static environments. The two bottom rows show dynamic environments.}
    \label{fig:exp_mp}
\end{figure*}

\subsection{7-axis Franka Robot Experiments}
We further conduct experiments on the Franka robot to demonstrate the effectiveness of CDF. 

We place several different obstacles such as spheres, walls, and rings, with randomly sampled initial and goal configurations. For a fair comparison, we use the neural representation for both CDF and SDF. Experiments are also repeated 100 times and we report the average results in Table~\ref{tab:MP_bench}. The CDF-based planners demonstrate better performance than SDF-based approaches in terms of success rate, tracking error and number of time steps. For QP controllers, our methods can run over 200Hz frequency, thanks to the high efficiency of neural networks.

For real-world experiments, we set up two different scenarios: static and dynamic environments. In the static environment, we place some obstacles in the robot workspace, such as blocks (for simple cases) and a shelf (for hard cases being highly non-convex). We use a RealSense D435 camera to capture the point cloud of the obstacles. For dynamic environments, we simply detect the obstacle according to the HSV color and convert it to point clouds. The same QP controller with IPOPT optimizer is used for motion planning. For static scenes, we wait for the QP controller to compute the full trajectory and then execute it on the robot. For dynamic scenes, we test the reactive motion generation and send the control commands online.
Qualitative results are shown in Fig.~\ref{fig:exp_mp}, showing the effectiveness of our approach.

\section{Discussion and Conclusion}
\label{sec:conclusion}

In this paper, we proposed to consider the geometry of robot configuration space as a distance field, and present a new representation called CDF to describe the topological structure of objects in configuration space. After discussing the formulation and properties, we introduced an efficient algorithm to compute and fuse CDFs. An implicit neural network encoding was also proposed to balance the accuracy, efficiency, and compression of CDF. We demonstrated the effectiveness of CDF with planar examples and with a 7-axis Franka robot in whole-body inverse kinematics and motion planning tasks.

CDF serves as a representation implicitly encoding inverse kinematics through a distance field. Both Cartesian grid and neural network representations facilitate the offline computation of inverse kinematics, thus enabling efficient online queries. In contrast, SDFs can be perceived as representations encoding the forward kinematics of a robot.

The pivotal step in CDF computation involves identifying zero-level-set configurations from a given point or object. Recent advancements in differentiable robot SDF representations have paved the way for efficient, accurate, and parallelizable robot SDF computation. This advancement ensures efficient solutions in optimization problems. The fusion strategy and neural network representation further enhance the computational efficiency of online queries.

CDF offers a global view of configuration space geometry. In practical applications, it is unnecessary to reconstruct the entire CDF, and selective querying of points of interest suffices. This adaptability enables the application of off-the-shelf techniques developed for task space to be extended to configuration space. Additionally, the combined use of SDFs and CDFs allows for a holistic understanding of geometry information across both task and configuration spaces, avoiding the need for calculating the nonlinear mapping.

Acknowledging the challenges posed by the high-dimensional configuration space and limited data, CDF computations encounter issues such as time consumption for offline computation, memory inefficiency for non-parametric representation, or compromised accuracy in neural network representation. The preference for neural representation is due to the trade-off between efficiency, compression, and accuracy, as well as the flexible structure that can be integrated into other frameworks. The learning results of neural networks are not fully optimized, as we mainly focused in this paper on the representation. Nevertheless, once a model is trained, the robot can re-use it multiple times later as a function approximator. 

\textcolor{black}{In addition to the balance between accuracy and efficiency, CDF currently has other limitations that need to be improved. First, the unsigned distance field is always non-negative, and using limited zero-level-set configuration samples can result in an inaccurate approximation for configurations close to the zero-level-set. It can be solved by computing the gradient of zero-level-set configuration samples and adding sign information to the CDF representation. Additionally, although CDF can be generalized to any points in the workspace and arbitrary configurations, the representation is based on the geometry and kinematics of the robot. It has to be computed separately for different robot models. The scalability of CDF to high-dimensional kinematic chains (e.g., full humanoids) also remains a challenge. Finally, the CDF measures the angular distance of joint configurations, which could be further extended to other distance metrics, such as the geodesic distance on the configuration space manifold, for better understanding the topology of the configuration space and benefiting manipulation planning tasks.} Future work will delve into exploring the broader applicability of CDF in various robotic problem domains, such as collision-free inverse kinematics, geometric motion planning, operation space control, multi-objective optimization, and robotic learning. 

\section{Acknowledgments}
This work was supported by the China Scholarship Council (No. 202204910113), 
and by the State Secretariat for Education, Research and Innovation in Switzerland for participation in the European Commission's Horizon Europe Program through the INTELLIMAN project (\url{https://intelliman-project.eu/}, HORIZON-CL4-Digital-Emerging Grant 101070136) and the SESTOSENSO project (\url{http://sestosenso.eu/}, HORIZON-CL4-Digital-Emerging Grant 101070310). 

\bibliographystyle{plainnat}
\bibliography{reference}

\newpage

\appendix
{\color{black}
\subsection{Differentiability of CDF}
The signed distance field is differentiable almost everywhere and satisfies the eikonal equation $\|\nabla f\|=1$. An illustration of the 1D signed distance field is shown in Figure \ref{fig:1dsdf}-a (left).

The CDF defined by Eq.~\eqref{eq:CDF} is an unsigned distance field. Although it still satisfies the eikonal equation, it is not differentiable at the zero-level-set (shown in Figure \ref{fig:1dsdf}-a (right)). Since the CDF is computed by finding the closest configuration point on the zero-level-set, it may also lead to non-differentiable points when the closest joint configuration changes (shown in Figure \ref{fig:1dsdf}-b). The CDF is thus differentiable in configuration space except at the isosurface (zero-level-set) samples and points where the closest joint configuration changes. Nevertheless, the left and right derivatives at those points are well-defined and satisfy the eikonal equation. Additionally, the non-differentiable interval is sparse in the configuration space, and we can use the subgradient instead if needed.

\begin{figure}[htbp]
    \centering
    \begin{tabular}{c}
        \includegraphics[width =0.95\linewidth]{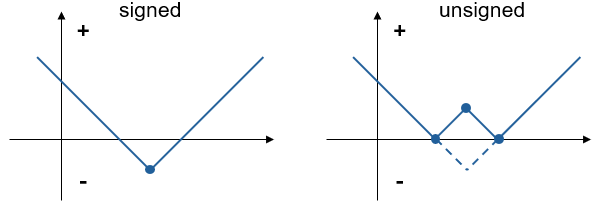} \\
        (a) \\
        \includegraphics[width =0.95\linewidth]{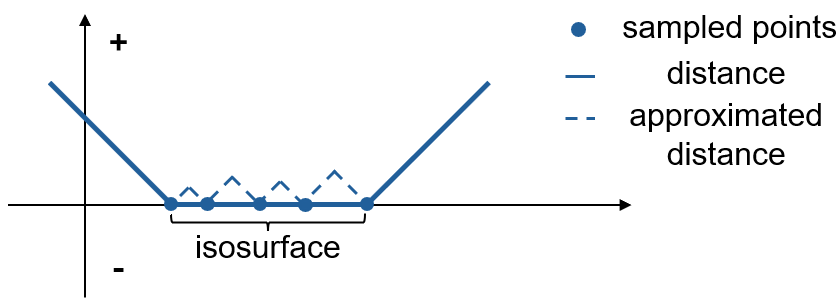} \\
        (b)
    \end{tabular}
    \caption{\textcolor{black}{Illustration of the differentiability of CDF.}}
    \label{fig:1dsdf}
\end{figure}

The neural CDF approximates the function using a multi-layer perceptron (MLP). The oscillation of training data near the isosurface may cause an inaccurate estimation of CDF and its gradient for inputs close to the zero-level-set. The differentiability of the neural CDF depends on the activation function of the network (for example, the ReLU function is not differentiable at zero) and the gradient can be analytically computed through backpropagation.

\subsection{Sensitivity to Noise}
In real-world scenarios, the observed data usually have noise and may affect the performance of CDF. To further evaluate the sensitivity of the neural CDF representation, we introduce Gaussian noise on input points with zero mean value ($\mu$) and standard deviation ($\sigma$) ranging from $0.01$ to $0.03$ and visualize the mean absolute error (MAE) and success rate (SR) in Figure~\ref{fig:lCDF_noise}. 
Although performance decreases with the increase of noise, the overall MAE and SR are still acceptable, particularly in scenarios with less noise ($\sigma=0.01$).

\begin{figure}[htbp]
    \centering
    \includegraphics[width =0.95\linewidth]{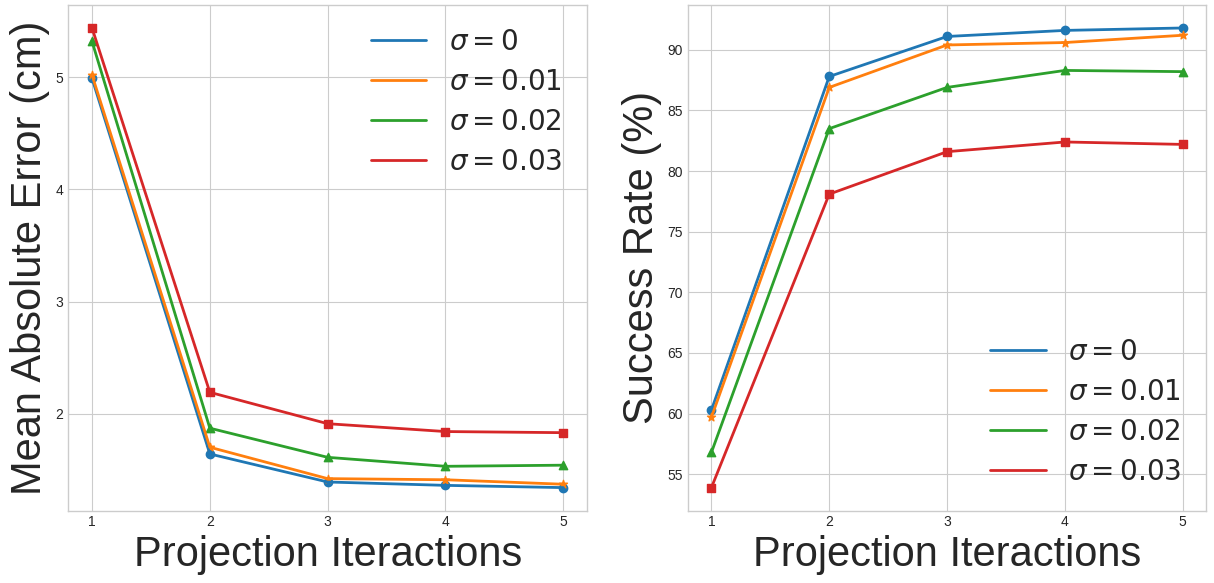}
    \caption{\textcolor{black}{The sensitivity of Neural CDF to Gaussian noise.}}
    \label{fig:lCDF_noise}
\end{figure}

}

\end{document}